\documentclass[sigconf]{acmart}
\AtBeginDocument{%
  \providecommand\BibTeX{{%
    \normalfont B\kern-0.5em{\scshape i\kern-0.25em b}\kern-0.8em\TeX}}}

\copyrightyear{2021}
\acmYear{2021}
\setcopyright{acmlicensed}\acmConference[CVMP '21]{European Conference on Visual Media Production}{December 6--7, 2021}{London, United Kingdom}
\acmBooktitle{European Conference on Visual Media Production (CVMP '21), December 6--7, 2021, London, United Kingdom}
\acmPrice{15.00}
\acmDOI{10.1145/3485441.3485648}
\acmISBN{978-1-4503-9094-1/21/12}

\usepackage{gensymb}

\begin{document}

\title{Automatic Camera Control and Directing with an Ultra-High-Definition Collaborative Recording System}

\author{Bram Vanherle}
\email{bram.vanherle@uhasselt.be}
\affiliation{%
  \institution{Hasselt University - tUL - Flanders Make, Expertise Center for Digital Media}
  \streetaddress{Wetenschapspark 2}
  \city{Diepenbeek}
  \country{Belgium}
  \postcode{3590}
}

\author{Tim Vervoort}
\email{tim.vervoort@uhasselt.be}
\affiliation{%
  \institution{Hasselt University - tUL - Flanders Make, Expertise Center for Digital Media}
  \streetaddress{Wetenschapspark 2}
  \city{Diepenbeek}
  \country{Belgium}
  \postcode{3590}
}
\author{Nick Michiels}
\email{nick.michiels@uhasselt.be}
\affiliation{%
  \institution{Hasselt University - tUL - Flanders Make, Expertise Center for Digital Media}
  \streetaddress{Wetenschapspark 2}
  \city{Diepenbeek}
  \country{Belgium}
  \postcode{3590}
}

\author{Philippe Bekaert}
\email{philippe.bekaert@uhasselt.be}
\affiliation{%
  \institution{Hasselt University - tUL - Flanders Make, Expertise Center for Digital Media}
  \streetaddress{Wetenschapspark 2}
  \city{Diepenbeek}
  \country{Belgium}
  \postcode{3590}
}


\begin{abstract}
  Capturing an event from multiple camera angles can give a viewer the most complete and interesting picture of that event. To be suitable for broadcasting, a human director needs to decide what to show at each point in time. This can become cumbersome with an increasing number of camera angles. The introduction of omnidirectional or wide-angle cameras has allowed for events to be captured more completely, making it even more difficult for the director to pick a good shot. In this paper, a system is presented that, given multiple ultra-high resolution video streams of an event, can generate a visually pleasing sequence of shots that manages to follow the relevant action of an event. Due to the algorithm being general purpose, it can be applied to most scenarios that feature humans. The proposed method allows for online processing when real-time broadcasting is required, as well as offline processing when the quality of the camera operation is the priority. Object detection is used to detect humans and other objects of interest in the input streams. Detected persons of interest, along with a set of rules based on cinematic conventions, are used to determine which video stream to show and what part of that stream is virtually framed. The user can provide a number of settings that determine how these rules are interpreted. The system is able to handle input from different wide-angle video streams by removing lens distortions. Using a user study it is shown, for a number of different scenarios, that the proposed automated director is able to capture an event with aesthetically pleasing video compositions and human-like shot switching behavior.
\end{abstract}

\ccsdesc[300]{Computing methodologies~Neural networks}
\ccsdesc[500]{Applied computing~Media arts}
\ccsdesc[300]{Computing methodologies~Graphics processors}
\ccsdesc[300]{Computing methodologies~Object detection}

\keywords{Video processing, automated directing, object tracking}

\begin{teaserfigure}
  \includegraphics[width=\textwidth]{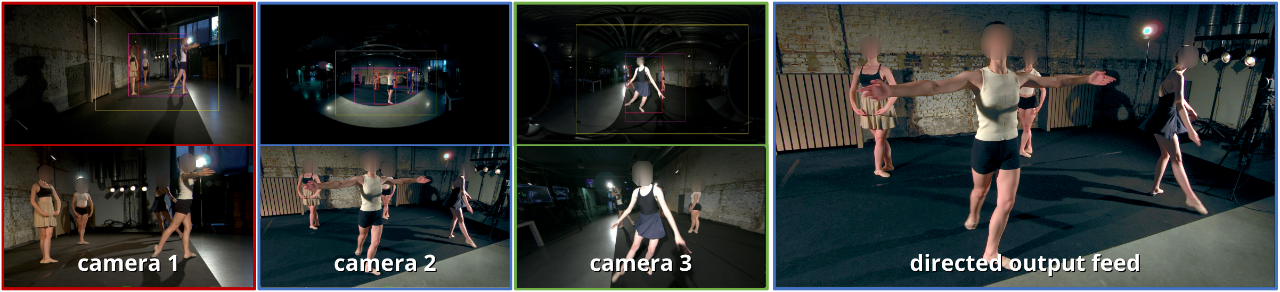}
\caption{Demonstration of how the collaborative recording system transforms three wide-angle input streams (top-left) into a directed output feed. The system uses Object detection to generate virtual framings for each camera (bottom-left).}
  \label{fig:teaser}
\end{teaserfigure}

\maketitle

\section{Introduction}
Due to the prevalence of devices that can stream multi-media, it is now easier than ever for people to stream and watch video. For event organisers, this means they can increase their attendance from thousands of physical people to millions of virtual attendees, as they can effortlessly distribute video of their events via the internet. With cameras becoming more advanced and cheaper, it is possible to capture the action from all angles. The end user however, often viewing on a small screen, does not want to view all angles. For the best experience, they want to see video from one angle, that is best positioned to observe the action.

Traditionally, a human director would select what camera feed should be shown to the viewer. Those cameras can either be fixed or pan-tilt-zoom (PTZ). Fixed cameras can only capture a part of the action, whereas PTZ cameras require an extra operator to point them towards the activity. To reduce the operational cost of having extra camera operators, whilst still being able to capture most of the event, a set of high field of view fixed cameras can be used. Using this type of camera, however, increases the difficulty for the director, as they now also need to select what region of the video feed to show the viewer.

In this paper, a system is proposed that allows event organisers to effortlessly and inexpensively create video content capturing their events in an exciting manner. The system consists of two parts: an ultra-high-definition and wide-angle acquisition system for broadcasting and a set of algorithms for automatically directing the final video content. The acquisition system is able to create virtual PTZ cameras that can move around in the wide-angle shots while removing lens distortions. Moving around multiple virtual cameras and selecting shots is a complex task, we therefore introduce an automated virtual camera operator and director. The proposed automated tools use existing deep learning models to detect people and objects in the video streams. These detections are used to generate virtual camera tracks, following interesting targets. A novel algorithm, based on cinematographic rules, determines at each point in time which virtual camera to select as output feed. It can be used for both live broadcasting and offline post production. The algorithm works fully automated, but can be tweaked by a number of parameters set beforehand.

To show the versatility of our method, we apply the system to a wide array of events, ranging from a basketball game, a dance performance, to a speaker on stage. To verify the effectiveness of our approach, a user study is conducted. Users of varying levels of expertise are asked what viewing angles they would prefer. These results are compared to the output of our algorithm.

\section{Related Work}
One of the first autonomous camera systems was introduced by \cite{intelligent_studios}. They developed a system that selects a framing for each camera based on a  number of vision algorithms. Which vision algorithm is used, is determined by interpreting the script. This approach is specific to cooking shows, and requires domain knowledge of the script beforehand. \cite{virtualCinematographer96} introduced a paradigm for generating camera positioning in a virtual environment. They introduce idioms that encode expertise about the scene to determine the camera shot type. Due to the system working in a virtual 3D scene, the algorithm can easily extract semantic information, reposition cameras and even reposition actors. \cite{virtual_lecture} created a system for automatically directing the video feed of a lecture. They use temporal differencing to determine what part of the screen to crop and use a bilateral filter to smooth movement. Their approach does not support more complex scenarios, multiple cameras or multiple actors. \cite{soccer_director} introduced an automatic production system for a more complex scenario, a soccer match. They use background subtraction to track the players and the ball. This information is used to recognize the game situation and select an appropriate area to crop. This method does not apply to other scenarios as they use hardcoded rules about soccer, also only one camera is supported. The works of \cite{basketball_director} and \cite{football_director} do support multiple cameras, but their methods only work for a specific sport, basketball and soccer respectively. \cite{camera_selection} introduced a system to present a human director with a selection of good camera angles for a soccer match. They do so by training an support vector machine on a number of features such as ball visibility and player distribution. The system does not propose a crop for high resolution imagery and requires human input. Pano2Vid~\cite{360_pano} is a automatic cinematography system for panoramic videos. It works by dividing the entire surface of the video into regions. From these regions \textit{glimpses} of 5 seconds are taken. A classifier is trained on human made videos to determine what an interesting glimpse looks like. Good glimpses are found throughout the video and a smooth camera track that passes through them is found. This system manages to show the viewer interesting parts of the video, but has no focus on generating visually pleasing sequences that resemble a human director. There is no multiple camera support either.

In this paper we introduce a system that can acquire ultra-high resolution wide-angle video from multiple viewpoints. As it is difficult to easily transform this type of imagery into a broadcast or video, an automated director method is developed. This is a general purpose method, that can handle all events in which humans play the central role. The algorithm can process high-resolution wide-angle video streams taken from multiple camera angles, and automatically transform them into an entertaining and natural looking video feed. 

\section{Acquisition system for ultra-high resolution wide-angle broadcast}
\label{studioone}
\begin{figure}[ht]
  \centering
  \includegraphics[width=\linewidth]{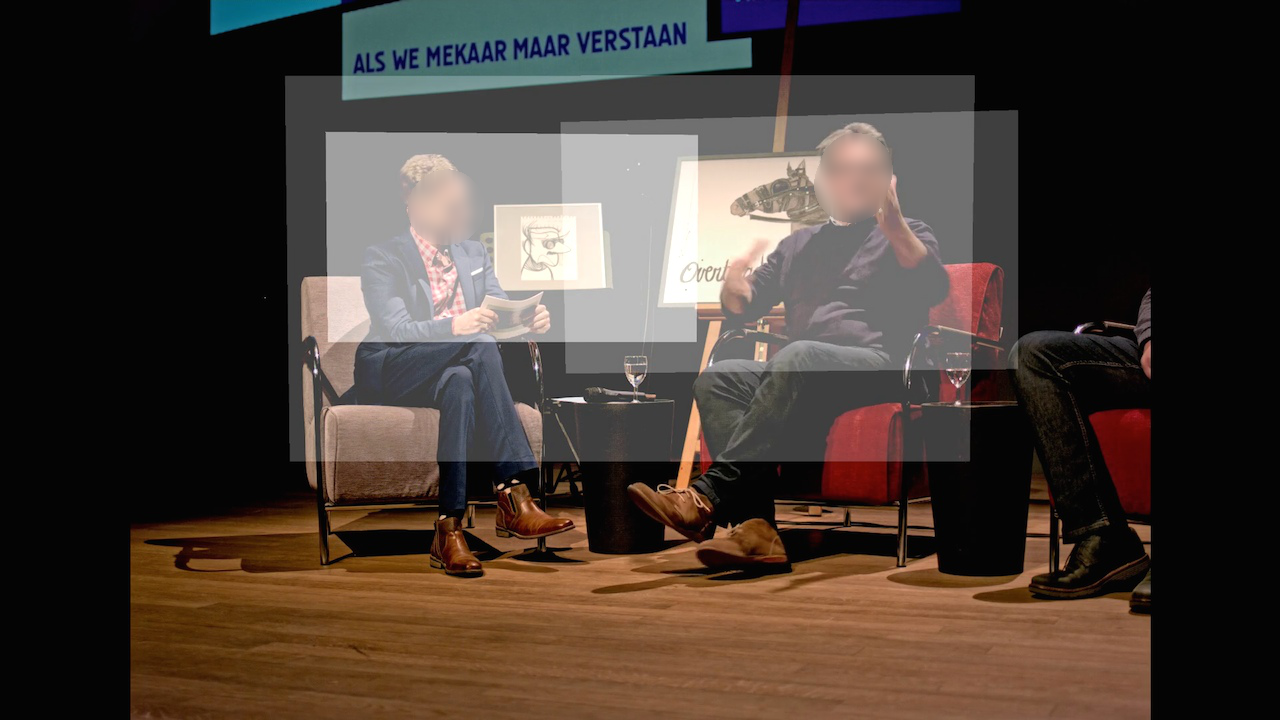}
  \caption{The acquisition used in a panel discussion. Here, the white rectangles show the virtual camera views; a close-up of every panellist and a group shot of the two guests.}
  \label{fig:studio_one}
\end{figure}

\begin{figure}[ht]
  \centering
  \includegraphics[width=\linewidth]{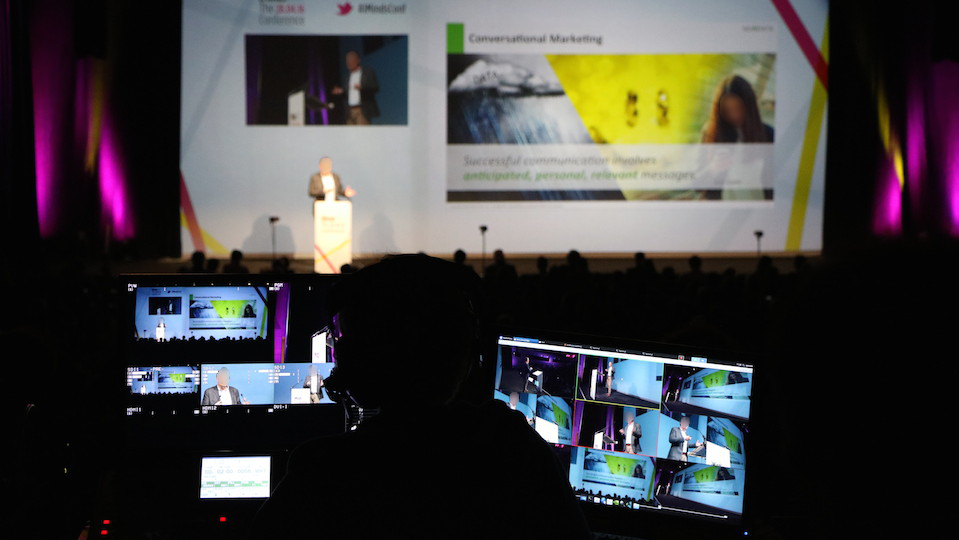}
  \caption{The acquisition system used at a conference, with four ultra-high-definition static cameras on stage and a broadcasting server in the front.}
  \label{fig:studio_one_back}
\end{figure}

To record the wide-angle video streams, required for collaborative recording, we have developed an in-house multi-camera video capture and production system. It enables recording and broadcasting with multiple ultra-high-definition (up until 8K resolution) cameras with wide-angle or omnidirectional lenses. The server computes multiple virtual cameras from any input camera in real-time. Like ePTZ cameras, the operator can virtually zoom in or out using a joystick to simulate camera operators. The software automatically converts the distorted wide-angle input into the correctly rectified output feeds. It can be used for both real-time broadcast and offline editing. Virtual camera movement, angle switching and color grading are stored as metadata and can be edited in post-production. A mobile set-up is shown in Figures~\ref{fig:studio_one} and \ref{fig:studio_one_back}. Using only a handful of compact cameras and a single server, the operator can single-handedly execute a multi-camera recording with high production value.

The benefit of using such a system is that only a single operator is required, and everything can be modified in post-production. This reduces the operational cost and makes multi-camera productions accessible to more organizations with lower budgets.

Because the camera zoom level and movements are entirely virtual, the system only records the full camera raw image. This way, multiple virtual camera angles can be computed from the same physical camera. This can be done in real-time or in post-production. In real-time, the operator can duplicate the camera to multiple virtual views. Then, each view can be controlled separately. During post-production, each camera can again be linked to one or more virtual views. For each view, a keyframe file is constructed that defines the virtual camera properties. This can be useful for group tracking. For example, for three people in a scene, the same camera can track the three people as a group, the three pairs of people, and the three individual people, resulting in seven possible virtual framings, all from the same input camera. This has the added benefit that fewer cameras are required compared to traditional multi-camera productions.

This system uses state-of-the-art video frame processing on the GPU, allowing it to handle multiple 4K or even 8K video streams in real-time.

We use this system as a backbone for our approach as it supports real-time high-resolution camera processing and conversions between different lens models. Combining the automatic directing algorithms, such as people tracking, cut-out creation, group tracking, smoothing, and camera angle switching that we propose in the rest of this paper, with such an acquisition setup will result in an autonomous broadcasting system.


\section{Collaborative Recording}
To be able to automatically direct the video feed of a certain event, a series of automatic rendering instructions is required. Concretely, the rendering instruction for the $i$-th frame is defined as $d_i = (s, c_x, c_y, z)$, with $s$ the index of the video stream to show, $(c_x, c_y)$ the center of the proposed framing and $z$ the factor by which the frame should be zoomed in. To generate such a set of instructions that manages to capture interesting events in a visually pleasing way, we propose the following pipeline.

Firstly, image data from the video streams are fed to an object detector whose task is to detect people and potential objects of interest in these images. Object tracking is used to transform these unordered bounding boxes into traces that belong to a specific person or object. The obtained traces can then be used to follow the persons or objects of interest over time. The camera tracking system uses these traces to generate smooth camera tracks that can be used in the final rendering. Finally, the automatic director decides which of the camera tracks should be shown, considering the user's settings. The director generates a series of rendering instructions that are interpreted to obtain the final image sequence. The following sections will detail more on each of these individual steps.

\subsection{Object Detection}
To generate an attractive sequence from a series of images, a high-level understanding of the scene is required. The events that our tool will focus on are human-centered, which creates a need to detect people's locations in the different camera feeds. Optionally, the user can specify other types of objects that would make the framing more interesting. These would then also need to be detected. Such objects could include a sports ball or a human face.

The humans and objects to detect can vary immensely in appearance, since it needs to work for a wide variety of scenes and contexts. When trained properly, Deep Neural Networks have the ability to generalize very well to unseen data. There is a plethora of Object Detection models available, but we opted for YOLOv4~\cite{bochkovskiy2020yolov4} as it is able to detect in real-time, while still achieving high accuracy. A model, pre-trained on the MS COCO~\cite{MSCOCO} dataset, is used since it includes the human class along with a large number of different object categories that could be used. 

In some situations, it is not of interest to include detections from every position on the screen. For example, in a sports game, we are not interested in the surrounding crowd of supporters. For this reason, it is possible to supply a binary mask for the object detection. The binary mask indicates what part of the image we want to include detections for. Bounding boxes that have no points inside this mask will be removed from consideration.

\subsection{Track Creation}
To be able to select interesting viewpoints and create meaningful camera trajectories from them, we need to track individuals and groups throughout the video frames. This is done by assigning the bounding boxes, obtained in the previous section, to individuals and by interpolating bounding boxes over missing frames. There are two types of track creation approaches: individual tracking and group tracking. 

\subsubsection{Tracking Individuals}
To track people throughout a video, each person's location is estimated at each timeframe using a Kalman filter~\cite{Kalman}. Based on these predictions, the bounding boxes found in that frame are assigned to a person using the Hungarian algorithm~\cite{hungarian}.

The track creation process is executed by iterating over each frame and keeping a list of Kalman filter instances for each object currently being tracked. A couple of steps occur for each frame. Firstly, for each of the presently tracked items, their respective Kalman filter instance is used to predict their bounding box's specifications for this frame. These predictions are then used along with the detected bounding boxes for this frame to construct a cost matrix $C$, where $C_{i,j}$ is the cost for matching the $i$-th predicted bounding box to the $j$-th actual bounding box. The cost is computed by the intersection over union between the predicted and the detected bounding box:
\begin{equation}
    IoU(P, D) = \frac{|P \cap D|}{|P \cup D|}
\end{equation}
with $P$ the predicted and $D$ the actual bounding box. Subsequently, the Hungarian algorithm is used to generate an optimal assignment between the detections and predictions. Since objects can enter or leave the scene at each frame, the Hungarian algorithm's assignments are not always correct. We therefore only consider assignments below a certain cost threshold as an actual match. Detections that were successfully matched to an active trace are used to correct that trace’s Kalman filter. For each active trace the predicted bounding box is added to that trace, as opposed to the detection. Storing the predicted bounding box allows us to smooth out the bounding boxes for the trace and fill in missing detections.

Detections for this frame that were not matched to any existing trace are used to create a new trace and Kalman filter instance. These are considered to be entering the view. Object traces to which no new detections have been added for a while are removed from the currently tracked objects and considered done.  Once this process is finished for each frame, all traces that are long enough and have a high enough percentage of frames for which there were actual detections are considered good traces. The tail of each trace is also stripped, starting from the last frame for which an actual detection was found.

\subsubsection{Tracking Groups}
If specified by the user, it is possible to create a track following all the people in the scene. This is done by firstly creating a new Kalman filter instance. The following process is repeated for each frame. The bounding boxes for all the individual traces for this frame, created in the previous section, are gathered. From these bounding boxes, the outliers are removed. This is done by computing the average center of all the bounding boxes and removing detections that are too far away from this center, using an adaptive threshold based on the size of the screen. The remaining bounding boxes are merged in an all-encompassing box, which is used to correct the Kalman filter. The prediction from the filter is added to the group track, again using the Kalman filter to both smooth out detections and fill in missing detections.

\subsection{Camera Tracks}
After all the traces of the bounding boxes for each person and group in the collection of input streams are generated, the next step is to create camera tracks that determine how the camera should move, to keep that specific entity in frame. These tracks will later be used to determine which track is the ideal choice for each frame and to generate the corresponding rendering instructions. The movement and the amount by which the camera zooms in or out at each frame should be smooth and continuous. For this reason, we cannot merely use the output of the Kalman filter as the trajectory for our camera track. Two different methods for determining the camera movement are presented. The first fits a spline trough a set of keypoints on the object trajectory to generate a smooth curve. Unfortunately, this can only be used in post processing. For this reason, a rolling average delayed smoothing filter is proposed to be used for online processing.

\subsubsection{Cubic Spline} \label{sec:cubic_spline}
To generate the desired camera tracks for a specific entity’s trace, we take the bounding box at a number of keypoints that are equally spaced in time. For each of these bounding boxes the center location $(x, y)$ and the zoom factor $z$ is taken into account. $z$ is how far the camera needs to zoom in to exactly fit the bounding box in screen and is computed as $z = min(\frac{s_h}{b_h}, \frac{s_w}{b_w})$ with $(s_h, s_w)$ the screen size and $(b_h, b_w)$ the bounding box size. Three cubic splines are then fitted through the series of $x$, $y$, and $z$ values at the selected keypoints, resulting in splines $S_x, S_y$ and $S_z$. Given a series of points $(x_i, y_i)$ such a spline $S(x)$ is a set of piece-wise cubic polynomials:
\begin{equation}
    S(x)=\left\{\begin{array}{cl}
    C_{1}(x), & x_{0} \leq x \leq x_{1} \\
    \cdots & \\
    C_{i}(x), & x_{i-1}<x \leq x_{i} \\
    \cdots & \\
    C_{n}(x), & x_{n-1}<x \leq x_{n}
    \end{array}\right.
\end{equation}
where each $C_{i}=a_{i}+b_{i} x+c_{i} x^{2}+d_{i} x^{3}\left(d_{i} \neq 0\right)$.

\begin{figure}[ht]
  \centering
  \includegraphics[width=\linewidth]{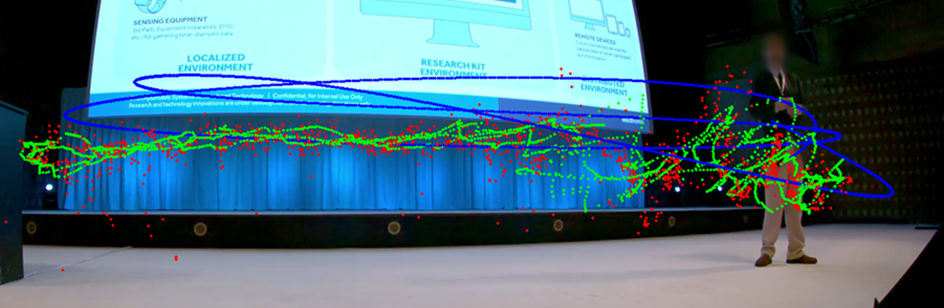}
  \caption{Illustration of the two types of filtering used to generate fluid camera tracks. The red dots represent the raw detected bounding box centers. The green dots are the filtered and interpolated positions using Kalman filtering. The blue lines are the fitted spline curves.}
  \label{fig:cubic_spline}
\end{figure}

It is now possible to generate a rendering instruction for video stream $s$ for an arbitrary frame number $n$ by evaluating these splines as follows:
\begin{equation}
    d_n = \{ S_x(n), S_y(n), S_z(n), s \}
\end{equation}
This creates a fluid motion for both the camera movement and zoom level whilst still following the desired entity. Figure~\ref{fig:cubic_spline} shows the effect of the extra smoothing provided by cubic interpolation. The red dots represent the original bounding box centers, the green dots are the positions after applying the Kalman filter, and the blue ones are the positions evaluated from the spline. The number of keypoints used to fit the cubic spline is a ratio of the total frames and is set as a parameter by the operator. Increasing the value for this parameter will results in the camera following the target more closely.

\subsubsection{Delayed Smoothing}
In a real-time environment, future positions of bounding boxes of detected objects are not available. Therefore, it is not possible to fit a curve to these non-existing coordinates. In such a situation, one can only use previous locations to predict future positions. Using previous coordinates of bounding boxes and a fixed time delay, the noise on the bounding boxes can be minimized, and new positions can be estimated.

The key idea is to use a FIFO queue that holds the consecutive bounding boxes of a detected object over time. The bounding boxes are not displayed immediately, just pushed to the queue. Once this buffer is full, the average bounding box $a$ is calculated over all boxes in the queue. Next, the delta between the previous position $p$ of the bounding box on the screen and the average bounding box $a$ is used to generate keyframes. Finally, the queue is cleared for new bounding boxes. The keyframes are used to transition smoothly from the last know position $p$ to the average position $a$ in the queue. This results in smooth movement and reduces noise on the bounding box positions.

The obvious downside of such an approach is that the delay is inherent to the simple algorithm. Fortunately, this delay can be overcome by compensating for this when the keyframes are generated. Adding a fixed offset to the keyframes hides the delay. For example, when a bounding box moves to the right over time, an offset can be added to the horizontal movement to overcome the delay.

Note that using a fixed-length buffer results in saddle points between intervals of length queue size. This saddle point can be smoothed by using an ease-in and ease-out function to make a more gradual transition. This ease-in and ease-out function can be simply implemented by varying the offset in a function of time. It begins by slowly rotating the movement from the previous series of points to the new series of points and then compensating for the rotation.

Delayed smoothing has excellent results for tracking people. Its behavior resembles human camera operators. A human operator anticipates movement and will gradually adjust the position, speed, and velocity of camera movements, just like delayed smoothing does.

\subsection{Automated Directing}
The last piece of the puzzle is to generate visually appealing cinematography for the input streams.  This is done by, for each frame, selecting which of the camera tracks, as obtained in the previous section, to show. Once this is known, the camera track information can be used to generate a cut-out at each frame, resulting in the final output video.

\subsubsection{Camera Track Selection} \label{sec:track_selection}
The result of the previous step is a pool of potential camera tracks, following different individuals or groups, from different camera angles, spanning different periods. The next goal is to select one of these to show for each frame in the final video. This is done, again, on a frame-by-frame basis. First, all of the eligible camera tracks for the particular frame are selected; these are the ones that contain the current frame number. The user can specify a specific object of interest that should be taken into account when selecting good camera tracks. This means that we prefer camera tracks that contain one of these objects; an example of this is a face. If this option is selected, the pool of eligible camera tracks is further decreased by removing the ones for which the framing does not contain the specified object. When the use of objects of interest is specified, we prefer to use tracks of individuals instead of tracks of groups since the group shot is likely to contain the object and so selecting it would not be of any interest. To incorporate this into our selection process, we choose the individual tracks that include objects from our eligible tracks, and if there were no such tracks, we would fall back to group tracks containing the objects. If no objects of interest are specified, the user can choose whether the algorithm should prefer group tracks or individual tracks. If the described process were to eliminate all camera tracks, we would select all eligible tracks for this frame.

We now have a select pool of camera track options for each frame. This needs to be further reduced to only one candidate. To do this, we score each option for each frame and select the candidate with the highest score. We have identified two metrics that help evaluate whether a shot is worth showing. The first is zooming out, this indicates that the target is moving towards the camera, which generally makes for an interesting shot. The value for zooming out is computed by the derivative of the inverse of the zooming spline computed in Section~\ref{sec:cubic_spline}. The second metric is the magnitude of movement. The motivation for this is that when a target is moving substantially, it should be interesting to look at. The movement score is the magnitude of the momentary movement vector. This vector is computed by taking the derivatives of the $x$ and $y$ splines. Since the movement value is indicated in pixels and the zooming value is a fraction of the screen, we cannot directly compare these. To solve this we multiply the zooming value by the height of the screen. The derivatives of these splines are used to find out the amount of change in these functions. Additionally, the use of these fitted splines makes our computations resistant to noise. This metric to use is specified by the user for each video stream and depends on preference and video type. This leads to the following function for the score:
\begin{equation}
    score(n) = \left\{
    \begin{array}{ll}
        s_w \cdot [S_z^{-1}(n)]' & \text{if scoring type is zoom} \\[5pt]
        \sqrt{S_x'(n)^2 + S_y'(n)^2} & \text{if scoring type is movement}
    \end{array}
    \right.
\end{equation}
For each frame the highest scoring camera track is selected to be shown. To avoid fast cuts that look unappealing, we enforce that each cut has a minimum duration. This is a parameter set by the user. In some scenarios this can give good results, but in other scenarios this can lead to the algorithm preferring mostly one camera. This is undesirable as it gets boring quickly. To avoid this, the user can also request to not always pick the best viewpoint. In this case the automated director will divide the timeline in segments with a random length between $l_{min}$ and $4 l_{min}$, with $l_{min}$ the minimal length of a cut, as set by the user. For each of these segments the highest scoring camera track over that period will be selected as final camera track. Unless it is the same as the previous segment, in that case the second best will be chosen.

Finally, we consider the scenario where there are no available tracks at a certain frame. This could be due to either no person of interest being on screen, or if the tracking system fails. The system handles this event by zooming out to show the entire screen, until a camera track is available again.

\subsubsection{Person Matching}
Automatic camera selection works adequately when applied to individual cameras or in use cases where no high level metadata of the person of interest is required, e.g., group tracking or single person tracking. In the case of multi-person broadcast, the director wants to cut from a particular person in one camera view towards the exact same person in a different view. He uses his intrinsic high-level knowledge of that person in both views. Doing this in an automated fashion is much more challenging. Our collaborative recording system has tackled this by integrating a person matching approach that relies on the deep person re-identification features of \cite{zhou2019omni}, where the deep convolutional neural network inputs an image of a person and returns a feature vector. For each of the camera tracks, a set of good frames is selected to compute the feature vector from. This is done by selecting all frames in which the bounding box of the tracked person does not intersect with any other bounding boxes. This will give the most descriptive feature, as no other people appear in the input. Furthermore, the frame is chosen for which the aspect ratio of the bounding box is closest to that of the neural network input. This avoids severe distortions by resizing the input. The full set of camera tracks across all camera feeds are then clustered together based on their mean feature vector using K-Means clustering. The amount of clusters is expected as a hyper parameter. The clusters of camera tracks will allow the system to generate automatic output feeds that are focused on specific designated persons, by simply applying the same camera track selection criteria as defined in the previous section, but now applied only on one cluster. In the case of uncertainty in person matching, the system falls back gracefully to a group shot. For example, in a dance performance with 3 artists, the system is able to export individual outputs for the entire group of dancers, as well as individual outputs for each dancer separately.

\subsubsection{Rendering Instructions}
Having chosen a camera track for each frame, we will now translate these to concrete rendering instructions. To this end, we divide our frames into shots; a shot is a sequence of consecutive frames for which the same camera track was chosen. In our system, we consider two types of shots: static and panning. A panning shot is a moving camera shot that simply follows the camera track. The rendering instructions for this type of shot are created by evaluating the $x$-, $y$- and zoom splines for each frame in the shot. A static shot does not move while always keeping the targeted entity on the screen. This shot's instruction is generated by merging all bounding boxes of this interval to an all-including bounding box. The center of this bounding box is then the center of the instruction, whereas the zoom is again the amount of zoom needed for the screen to contain the box.

The user can also specify whether camera tracks from any video should focus on the entire person or their face. If a person as the whole is selected, nothing changes, if the user prefers to focus on the face, the zoom factor is slightly increased, and the $y$ value of the rendering instructions is moved upwards by a quarter of the height of the bounding box.

Additionally, the user can specify a zoom factor per input stream. This derived zoom value $z$ is multiplied by this factor before generating the final crop. This gives the user more control over how the shots from this camera look, allowing them to emulate different types of cinematic shots, such as: wide, medium and close-up shots. Future work could consider how the best type of shot can automatically be derived from the video content.

\subsubsection{Lens Distortion}
\begin{figure}[b]
  \centering
  \includegraphics[width=\linewidth]{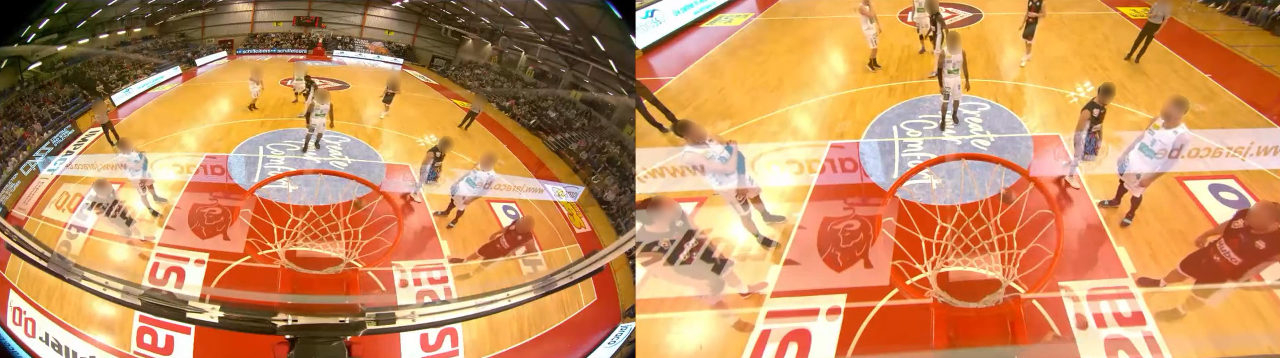}
  \caption{Camera cut-outs. (left) cropping in the wide-angle or equirectangular input feeds result in incorrect distortions. (right) re-projecting the input frame towards the correct rectified output camera model.}
  \label{fig:basket_studio_one}
\end{figure}
Since the proposed method can handle both wide-angle and omnidirectional video streams, the final step will be to convert the equirectangular format to a standard camera model. The pipeline described above can be coupled to our in-house system described in section~\ref{studioone}. The offline approach described above can analyze a recording made by the system and generate the corresponding render instructions that can be played back by the system in real-time. Here, 2D pixel coordinates of the center of the bounding box are converted to angles that in turn rotate the virtual eye in the system. This conversion is done using the built-in camera calibration settings and transforms a 2D coordinate to the correct 3D world coordinate system using Euler angles~\cite{euler}.

In an online approach, object detection and camera switching are plug-ins for the software. It downloads the camera frames from the GPU, detects people in the frame, generates bounding boxes, and chooses a virtual angle as a viewpoint. The size of the bounding box is converted to a field of view angle that the system can interpret. When cropping out the virtual camera on the original raw wide-angle or equirectangular video, the output feed will not be physically correct. This is shown in Figure~\ref{fig:basket_studio_one} on the left. In the right figure, we apply a real-time transformation of the same viewpoint to a more plausible camera lens model.

\subsection{User Parameters}
As mentioned throughout this section, the user needs to specify a number of parameters that control the behavior of the automatic director. An overview of all possible parameters is given below.
\subsubsection*{Per video settings}
\begin{itemize}
    \item \textit{Group or individual tracks}: Specify whether individuals and/or groups should be tracked.
    \item \textit{Prefer individual tracks} If set to true, individual tracks are preferred over group tracks, otherwise the other way around.
    \item \textit{Zoom type} Determine if the camera should zoom on the full body or on the upper body.
    \item \textit{Camera type} Determine if the camera should pan or is in a fixed position.
    \item \textit{Zoom factor} A float specifying how much the camera should zoom in.
    \item \textit{Fitting} A float determining how closely the camera follows the target, i.e., how many keypoints are used when generating the spline.
\end{itemize}

\subsubsection*{Director settings}
\begin{itemize}
    \item \textit{Minimal cut length} Specify the minimum amount of seconds the director should wait to switch shots.
    \item \textit{Best viewpoint always} Should the director always pick the best viewpoint, or can they switch to have more intricate directing.
\end{itemize}

\section{Results}
\begin{figure}[b]
  \centering
  \includegraphics[width=\linewidth]{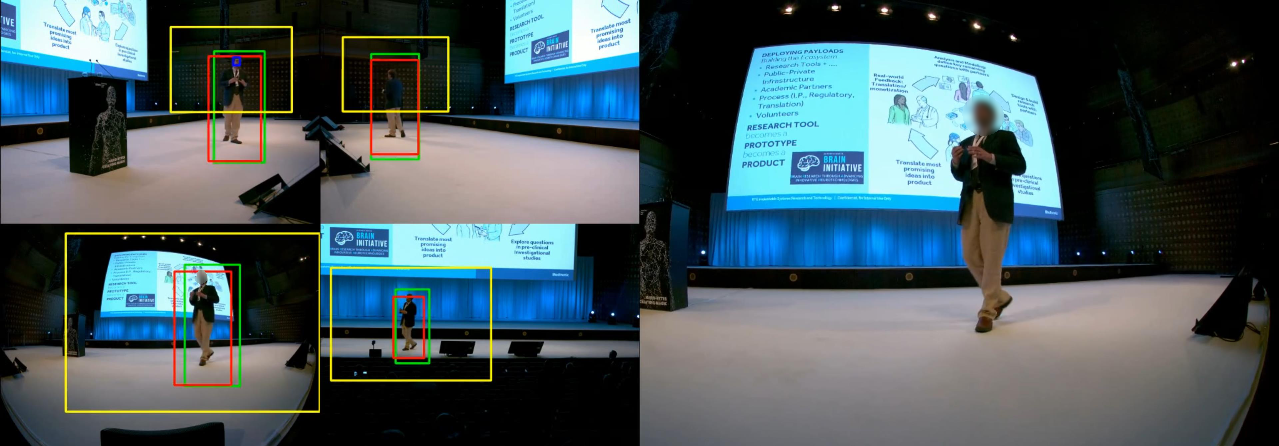}
  \caption{PoC 1 - single speaker on stage. (left) the four input feeds with camera cut-out detections. (right) the directed output feed.}
  \label{fig:test_conference}
\end{figure}

We evaluated the system with four different proof of concept (PoC) demonstrators with increasing complexity. The four different proof of concepts have been recorded with one to four ultra-high definition (4K and 8K) cameras. The directed output feed still retains a minimum of high-definition output resolution. The first proof of concept is to automatically broadcast a single speaker on stage at a conference, shown in Figure~\ref{fig:test_conference}. Here, four wide-angle cameras are used to record the event.

\begin{figure}[t]
  \centering
  \includegraphics[width=\linewidth]{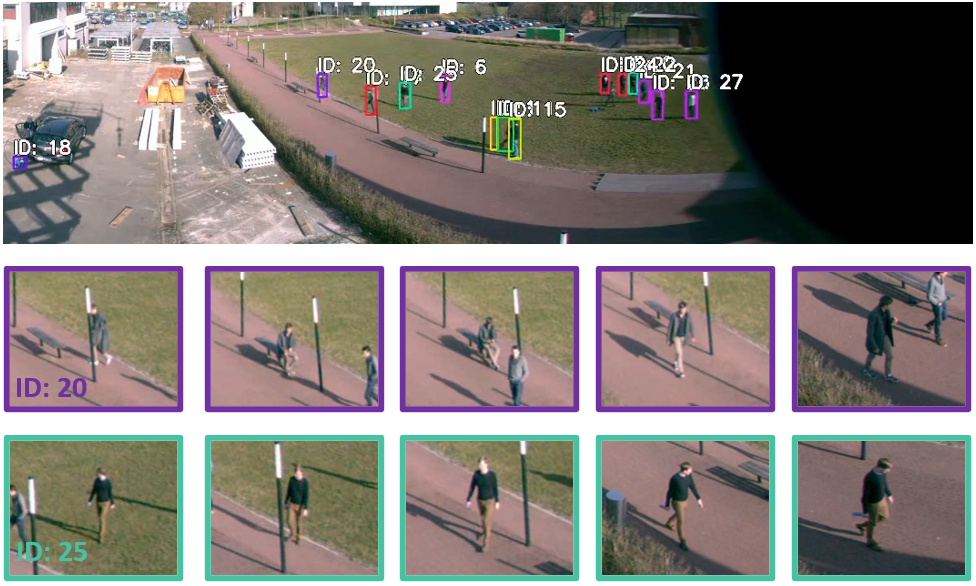}
  \caption{PoC 2 - Multi-person surveillance. Each detected individual is assigned a unique ID and tracked over time. The collaborative recording tool can decide who to follow based on this assigned ID (e.g., here cut-out videos are shown for two persons, with respective IDs 20 and 25).}
  \label{fig:multiple_people}
\end{figure}

The second proof of concept is a traffic surveillance environment where an ultra-high resolution omnidirectional traffic camera is used to simultaneously track numerous people walking around a campus. For example, in Figure~\ref{fig:multiple_people}, we can simultaneously create a virtual output feed for person with id 20 and 25 while still capturing the entire traffic environment.

\begin{figure}[t]
  \centering
  \includegraphics[width=\linewidth]{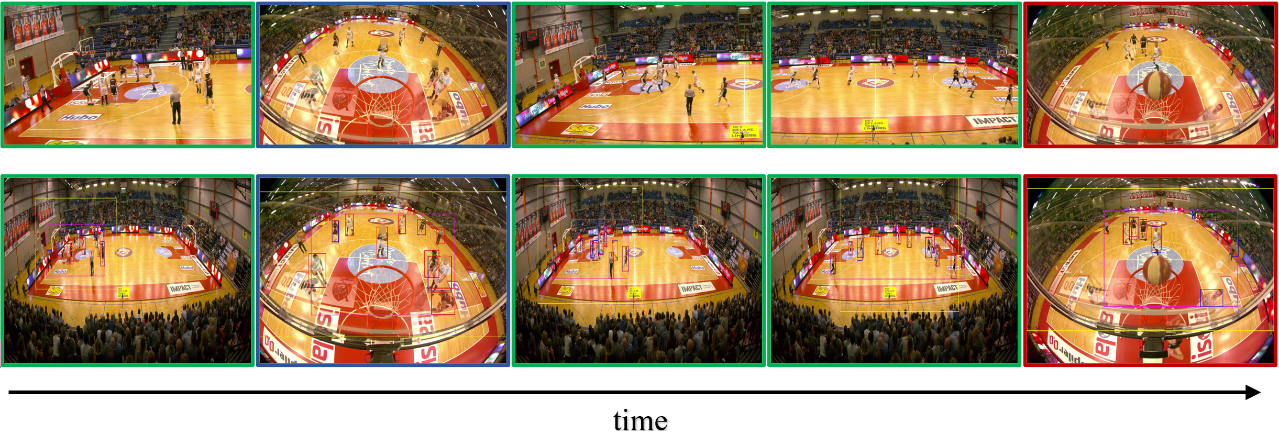}
  \caption{PoC 3 - Collaborative recording of a basketball game. (top) output sequence of the directed result (bottom) raw inputs from camera with bounding box detections and cut-out visualizations. The video sequence is played from left to right. The setup consists of three cameras, color coded with blue (left basket), green (side view) and red (right basket). Best viewed digitally. }
  \label{fig:test_basket}
\end{figure}

In a third proof of concept, we recorded a basketball game where we focused on creating group framings and following fast human movement. This game was captured using an 8K camera on the side and two 4K cameras, one above each ring. The results are shown in Figure~\ref{fig:test_basket}. The results over time are shown from left to right. The top part of the figure shows the output directed video. The bottom part are the selected raw input streams, illustrating the different group detections. The three different cameras are color coded with blue (left basket), green (side view) and red (right basket).

\begin{figure}[t]
  \centering
  \includegraphics[width=\linewidth]{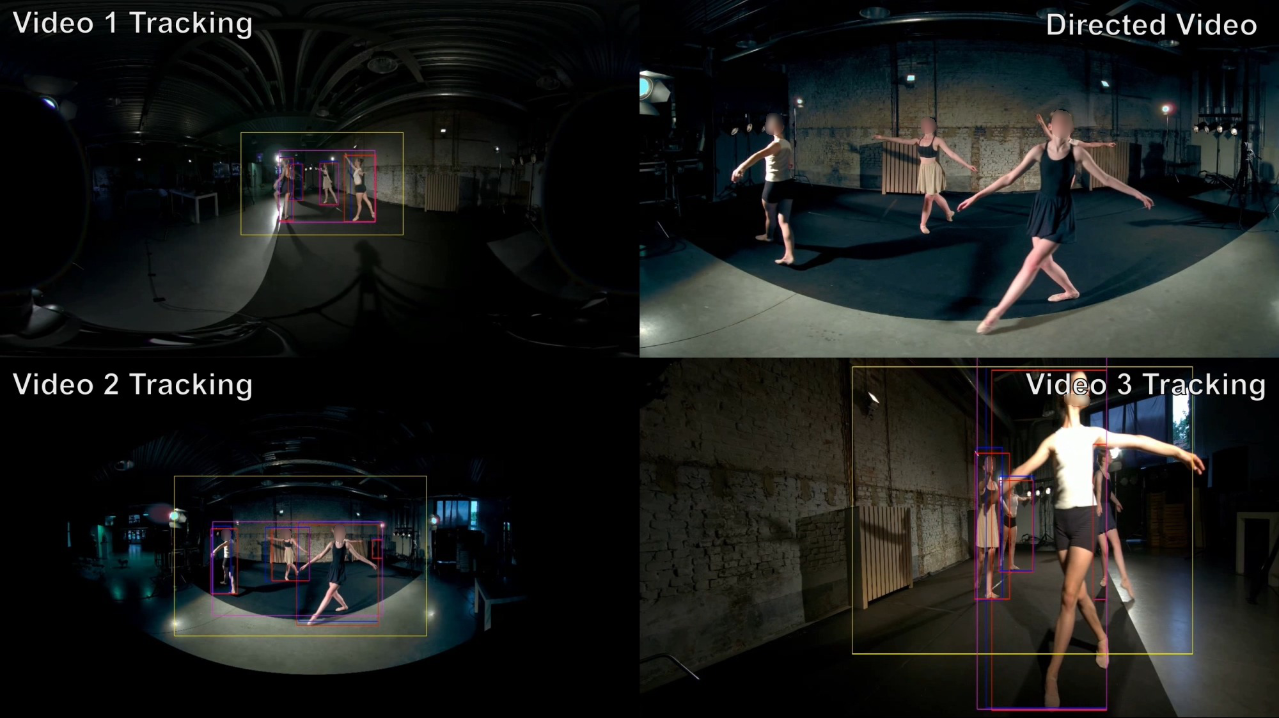}
  \caption{PoC 4 - collaborative recording of a dance performance, captured with 3 ultra-high definition cameras.}
  \label{fig:test_dance}
\end{figure}

\begin{figure}[t]
  \centering
  \includegraphics[width=\linewidth]{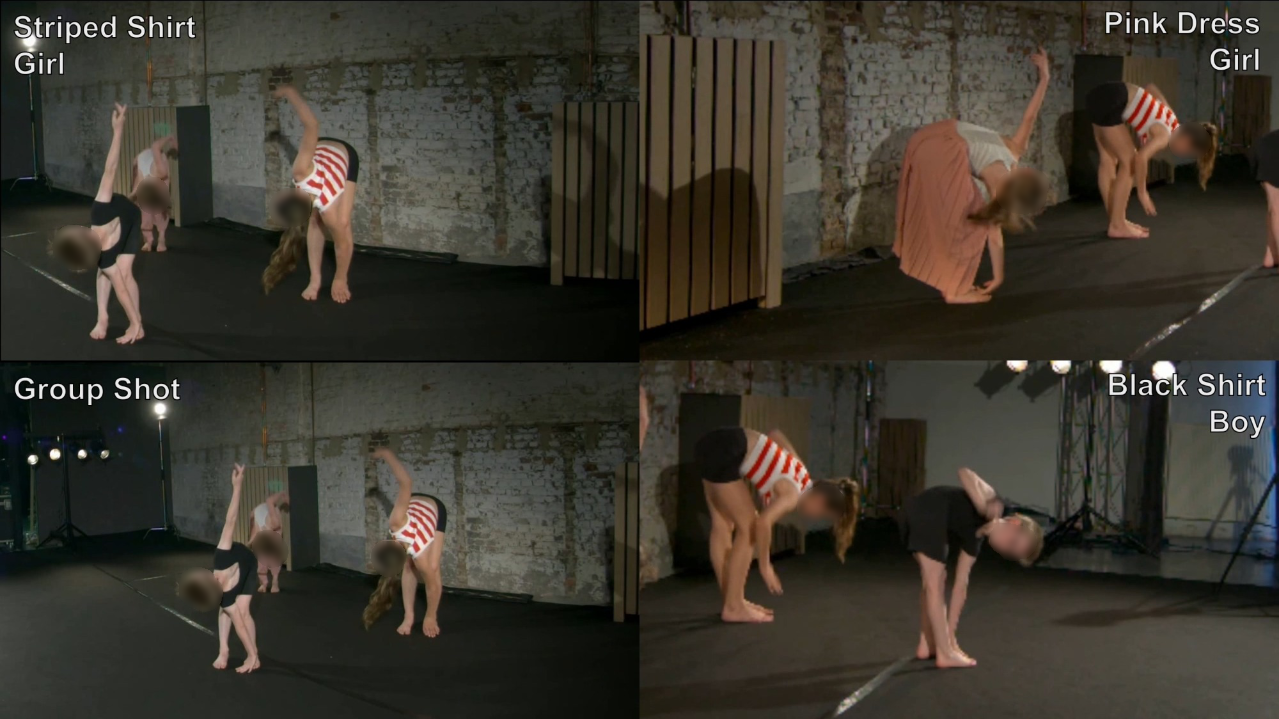}
  \caption{Create targeted output videos for individual persons using person re-identification.}
  \label{fig:dance_reid}
\end{figure}

In the fourth and last proof of concept, we captured a group of four dancers doing a choreography. In this example, both group shots and individual shots are incorporated. The operator has the freedom of choice to track individual performers or the full group. A single frame of the resulting video is shown in Figure~\ref{fig:test_dance}. Individual shots versus group shots are shown in Figure~\ref{fig:dance_reid}. The quality of the output feed is highly correlated with the quality of the individual person detection and re-identification. We have observed that the person re-identification is not flawless and can fail in certain circumstances, e.g., overlapping bounding boxes, bad lighting, etc. To still retain decent output feeds in such events, we gracefully degrade towards group tracking when the confidence for the individual tracking falls below a certain threshold.

\begin{figure}[b]
  \centering
  \includegraphics[width=\linewidth]{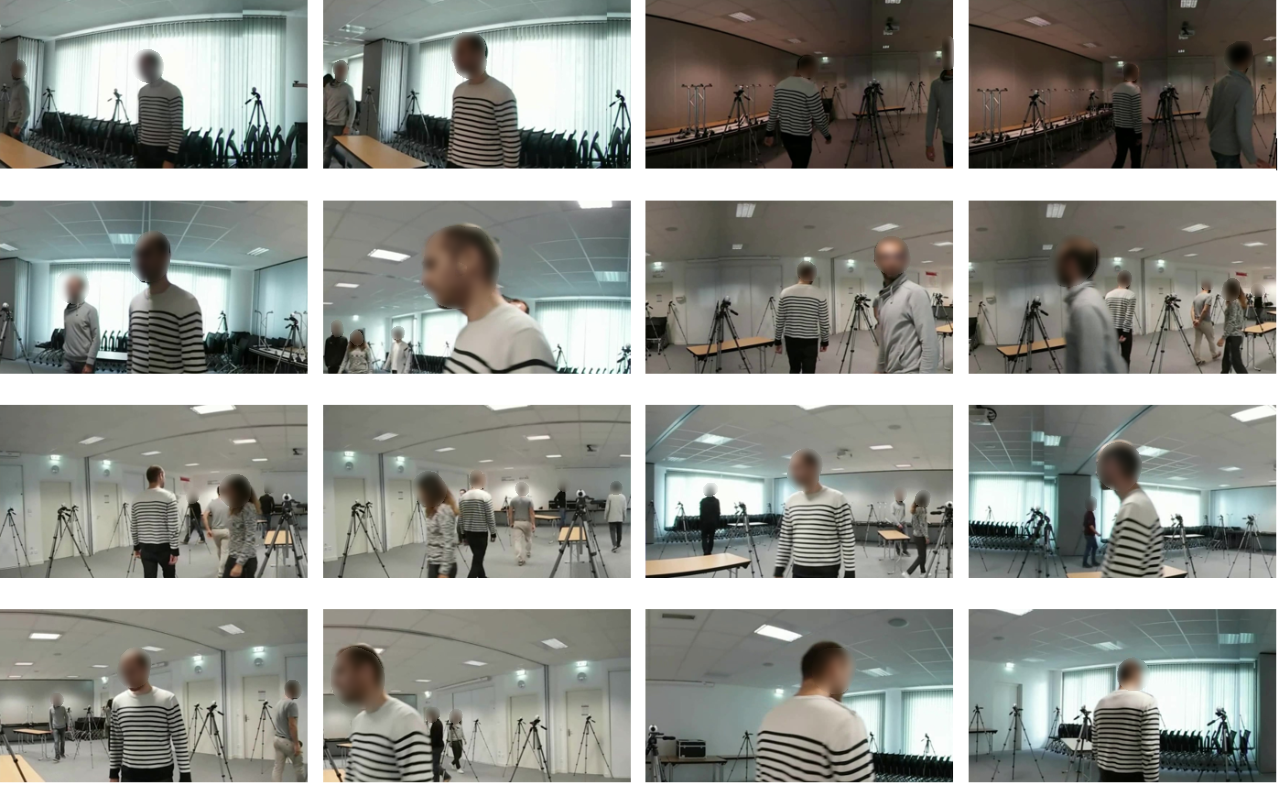}
  \caption{Automated direction and camera cut-outs on free viewpoint video. Tracking of the person in the white sweater using person re-identification across 40 free viewpoint cameras (indoor walk dataset, courtesy of \cite{freeviewpointvideo}).  The frames can be interpreted from top left to bottom right with 1 seconds interval in the video.}
  \label{fig:test_reid}
\end{figure}
To test our person matching and person re-identification on a more challenging dataset, we evaluated it on the online 360 degree free viewpoint dataset of \cite{freeviewpointvideo}. The dataset consists of 40 omnidirectional cameras, evenly spaced in an indoor environment, where multiple persons are walking around freely. We have first transformed all cameras into the equirectangular format and executed the person tracking algorithms. Then we generated re-identification features for each tracked person in each camera and try to identify similar persons across views. To decide if and when to cut to a different camera view, we applied the same high-level directing rules of person track selection (Section~\ref{sec:track_selection}). For example, the view with the largest bounding box is a good candidate, or when the bounding box of the person at interest is growing. Figure~\ref{fig:test_reid} show results of an automated broadcast of one selected individual. In this example we selected the man with the white sweater and tried to follow him across the room.  The results are quite promising, overall the person continues to be in frame, however the results are far from perfect. Occasionally the system switches to a wrong person because the re-id features are not always unique, e.g., two different persons with the same color of pants and sweaters are hard to distinguish. Furthermore, the rules for viewpoint selection are still relatively simple and need to be extended in the future with more realistic rules to result in a better viewpoint selection. For example, in this result, the viewpoint selection does not make a distinction between front facing and back facing angles.

\section{Evaluation}
To evaluate the cutting behavior of our collaborative recording system, we compared the output of the system in offline processing mode with the edits of multiple users. This experiment is a suggestion on how we can assess a system that switches between camera angles. Our experiment asked professional video editors and non-editors to edit a short fragment of our basketball game and three dance recordings. Then, the edits were compared based on cutting frequency, timing, and shot selection.

We created a web interface that shows the user a broadcast output and a multi-view of three camera angles. He or she can switch between camera angles by clicking the tiles of each camera in the multi-view. The videos play in sync, and the user switches between camera angles, the software logs every cut. It stores the current time in the video in seconds and the camera angle. We also created an array of cuts for the export of our collaborative recording system. Then these two arrays can be compared.

To compare the cutting frequency and timing, we calculate a root mean square error (RMSE) value between the current cut of one array and the nearest corresponding cut of the other array. Let's name the array of the user cuts $a$, and the array of the system cuts $b$. For every cut in $a$, we will search for the nearest cut in $b$. To find the nearest cut, we calculate the difference $d$ between the cut times for the cut in $a$ and the cut in $b$. The nearest cut is the one with the smallest difference $d$. When we have our nearest corresponding cut, we use difference $d$ in the RMSE calculation. The normalized RMSE here is the square root of the sum of quadratic differences $d$ for every cut in $a$, divided by the length of $a$. The calculation is shown in Formula \ref{eqn:rmse} where $l$ is the length of array $a$.
\begin{equation}
\label{eqn:rmse}
RMSE = \sqrt{\frac{\sum_{i=1}^{l} d_{i}^2}{l}}
\end{equation}
Next, we evaluate the shot selection. We calculate the overlap of identical shots over time, i.e., when both the user and our system selected the same shot $s$ over some time. We can calculate the overlap fraction based on our array of cuts by slightly modifying the list. For every cut, we also calculate the end time. The end time for every cut is set as the start time of the next cut. For the last cut in the list, the end time is the duration of the video. The overlap is calculated for every pair of cuts. For every cut in $a$ and every cut in $b$, we calculate the overlap between them. The sum of all the overlaps is the total overlap of both edited videos. We divide this number by the length of the video and obtain the overlap percentage.

The function to calculate the overlap between two cuts is defined as follows: take the minimum $m$ of the end time of the cut in $a$ and the cut in $b$, take the maximum $n$ of the cut time of the cut in $a$ and the cut in $b$ and subtract maximum $n$ from the minimum $n$. When the result $r$ is negative or zero, the cuts do not overlap. When the result $r$ is positive, the two cuts overlap for a duration of length $r$. This calculation is illustrated in Formula \ref{eqn:overlap}.
\begin{equation}
    overlap = \frac{\sum_{i=1}^{l} (min(a_{i_{end}}, b_{i_{end}}) - max(a_{i_{time}}, b_{i_{time}}))}{duration(a)}
    \label{eqn:overlap}
\end{equation}
A final measure is the F-score. By using precision and recall, we measure the average F-score for every possible camera angle between two arrays of cuts. The F-score for every angle $a$ is calculated by for every frame $f$ comparing the visible camera angle. If both angles are $a$, the frame $f$ is counted as a true positive. If both angles are not $a$, the frame $f$ is counted as a true negative. If the first angle is $a$, but the second is not, it is counted as a false positive. If the first angle is not $a$, but the second angle is, it is counted as a false negative.

Precision is calculated by dividing the number of true positives by the sum of the number of true positives and false positives. Recall is calculated by dividing the number of true positives by the sum of the number of true positives and false negatives.
The final f1-score is calculated by Formula \ref{eqn:f1}. We calculate this f1-score for every camera angle.
\begin{equation}
    f1 = \frac{(2 \times precision \times recall)}{(precision + recall)}
    \label{eqn:f1}
\end{equation}
Before the user could start editing, we asked a single question: 'How would you rate your experience with multi-camera editing?'. The user may choose between: 'No experience', 'Some experience', 'A lot of experience'. These questions are used to filter results. As such, professionals can be compared with other professionals, non-professionals, and our system.

\subsection{Evaluation Results}
We asked five professional editors and nine novice users to edit the basketball and dance videos using the tool described above.

In Table~\ref{tab:clip_metrics} some metrics pertaining to clip length and angle selection are shown. A first general observation is that our automated system makes more cuts than a human editor for the basketball demo. Hence the average clip length for the system is shorter than the average clip length for users. For the three dances, the average clip length and the number of cuts is almost identical. Furthermore, we see that the shot frequency (the time a single camera angle is used during the whole montage) is inline between a user and the system for the three dances. In the basketball demo, our users preferred the camera angles at the baskets above the panning center camera, where the system has a preference for the center camera. In addition, in our test, the users with more experience cut slower than users with less experience; this is true for all demos.

In Table~\ref{tab:shot_metrics} the overlap, RMSE and F-score metrics are used to compare the editing behaviour of users to that of other users, to that of the system and to themselves. A first observation is that there is not a large amount of overlap between the cuts of the user and the system, ranging between 54\% and 34\% for all scenarios. This seems to indicate that the system does not show human-like behaviour, but when investigating the average overlap between users among themselves, we see that this is also in the 55\% to 45\% range. This indicates that among humans there also is no consensus for what a good edit is. The RSME is for most scenarios lower between the average user and the system than between the users themselves. This means that the points in time when the system decides to cut to another angle closely resembles the cut moments chosen by the users.

Furthermore, a single un-experienced user resembles more to another un-experienced user than to the system. However, an experienced user resembles equally as much to the system as to other experienced users. We can infer that an experienced editor: a) cuts slower and b) has a unique style that is not matched by either other editors or the automated system. So far, the system resembles a human editor with little or no experience better than an editor with more experience.

Some users also did multiple runs. This allows us to compare an edit session of a single user with himself. The results are not surprising. Naturally, a user resembles himself more than the system or another user. However, no one did multiple runs that match entirely. Shot overlaps between multiple runs never reach 80\% while shot overlaps between different users or a user versus our system never reach 70\%.

\begin{table}[]
\scriptsize
\caption[evaluation-statistics]{Metrics comparing the editing behaviour of the average user to that of system.}
\begin{tabular}{lllllllll}
                    & \multicolumn{2}{c}{Basketball} & \multicolumn{2}{c}{Dance 1} & \multicolumn{2}{c}{Dance 2} & \multicolumn{2}{c}{Dance 3} \\
                     \cmidrule(l){2-3}  \cmidrule(l){4-5}  \cmidrule(l){6-7}  \cmidrule(l){8-9}
                    & Users        & System      & Users       & System       & Users        & System  & Users  & System  \\
Clip length  & 6.4s       & 3.7s      & 17.2s      & 11.5s      & 20.3s      & 13.7s      & 14.4s       & 12.8s      \\
Cut count   & 11.4        & 19          & 6.8         & 8.0            & 6.8    & 8& 21.6     & 21           \\
Angle 1     & 34\%       & 17\%      & 12\%      & 14\%      & 14\%      & 27\%      & 45\%      & 54\%      \\
Angle 2     & 46\%      & 70\%     & 59\%      & 47\%      & 52\%      & 40\%         & 42\%      & 40\%         \\
Angle 3     & 21\%      & 13\%     & 28\%      & 39\%      & 34\%      & 22\%      & 13\%      & 6\%      
\end{tabular}
 \label{tab:clip_metrics}
\end{table}

\begin{table*}[]
\footnotesize
\caption[evaluation-users]{Metrics computed to analyze overlap in cuts and shot selection behaviour. Users are compared to other users, the system and themselves.}
\begin{tabular}{llllllllll} 
 & \multicolumn{3}{c}{Basketball} & \multicolumn{2}{c}{Dance 1}  & \multicolumn{2}{c}{Dance 2}  & \multicolumn{2}{c}{Dance 3} \\
 \cmidrule(l){2-4} \cmidrule(l){5-6} \cmidrule(l){7-8} \cmidrule(l){9-10}
Metric             & versus user & versus system & versus himself & versus user & versus system & versus user & versus system & versus user  & versus system\\
\cmidrule(l){1-10}
Overlap all angles & 55.31\%            & 53.78\%        & 72.95\%     & 49.55\%            & 35.55\%  & 48.7\%             & 33.79\%  & 44.85\%            & 42.26\% \\
Overlap angle 1    & 29.76\%            & 20.51\%        & 47.26\%     & 18.17\%            & 0.32\%   & 15.45\%            & 8.43\%   & 34.88\%            & 31.69\% \\
Overlap angle 2    & 0\%                & 54.14\%        & 58.33\%     & 45.8\%             & 29.57\%  & 43.81\%            & 28.97\%  & 29.53\%            & 23.81\% \\
Overlap angle 3    & 18.56\%            & 18.93\%        & 49.35\%     & 16.37\%            & 18.62\%  & 23.11\%            & 15.6\%   & 5.96\%             & 12.11\% \\ \midrule
RMSE all angles    & 3.06               & 1.93           & 1.4         & 2.59               & 1.62     & 2.59               & 1.62     & 1.5                & 1.72    \\ \midrule
F-score angle 1    & 46.43\%            & 34.54\%        & 59.93\%     & 29.51\%            & 0.6\%    & 25.14\%            & 12.2\%   & 51.17\%            & 38.33\% \\
F-score angle 2    & 59.04\%            & 63.61\%        & 72.01\%     & 61.49\%            & 45.02\%  & 59.7\%             & 44.29\%  & 44.01\%            & 37.76\% \\
F-score angle 3    & 29.98\%            & 30.88\%        & 60.17\%     & 25.69\%            & 29.32\%  & 33.83\%            & 25.26\%  & 10.33\%            & 13.69\% \\
\end{tabular}
 \label{tab:shot_metrics}
\end{table*}

Using this simple evaluation method we can conclude that our system is human-like but resembles a novice editor. Mainly, the system cuts faster than humans and favors the center panning camera much more. The discrepancies seen in the results may be linked to system parameters such as threshold values (minimum shot duration and maximum shot duration) and the cost function used to select a new camera angle (e.g., prefer a fixed camera over a pan-tilt-zoom camera).

\section{Conclusion and Future Work}
In this article a method has been introduced in which many state-of-the-art techniques are combined to assist a video production team for broadcast. To validate this approach, a user study was conducted. Our experiments have shown that everyone edits the same sequence differently, highlighting that camera operation and directing for production are entirely subjective. It is not necessarily our goal to replace human camera operators and human directors. We believe that this system can be a great tool to help the operators and to save time and human resources by giving good framing and camera angle suggestions.\\
In our offline analysis, all frames must be processed by an object detector. This network does not work in real-time for ultra-high resolutions. Hence, it takes quite some time to analyze multiple camera angles and generate traces. This process could be sped up by using faster object detection and person re-identification algorithms.

It would be even better to incorporate all algorithms in a real-time system. Then, it can be used on location and used in real-time applications. We have already tested online object detection but have not created a real-time person re-identification system.

Next, if the user could change parameters, the system's behavior can be tailored to everyone's needs. Right now, the user can already specify if a camera should be static or pan and tilt to follow people. Some parameters that could tailor the experience can be cut frequency, a weighted cost function that favors camera angles over others, movement over static, and others. Furthermore, more rules for viewpoint selection could be implemented, resulting in a more diverse production.

Finally, the result of our system is dependant on the quality of object detections, smoothing, and person re-identification. Now, person detection is not quite perfect and results in noisy bounding boxes. That is why smoothing is needed. However, if these detections become better over time, the tracking will also be better, and virtual camera movement will mimic human camera operators. Also, when person re-identification improves, the group tracking and person tracking will be more consistent.

We are convinced that the system described in this article can be used in a variety of applications. The results described in this paper show that the different proposed technologies can be combined into a working proof of concept applicable for different use cases.

\begin{acks}
This work is partially supported by the FWO (SBO Project Omnidrone [S003817N]) and the Special Research Fund (BOF) of Hasselt University (Mandate ID: BOF20OWB24).
\end{acks}
\bibliographystyle{ACM-Reference-Format}
\bibliography{references}

\appendix
\end{document}